\documentclass[authorversion,nonacm]{acmart}
\pdfoutput=1
\usepackage{tikz}
\usetikzlibrary{fadings}
\usetikzlibrary{shapes}
\usetikzlibrary{shapes.arrows}
\usetikzlibrary{shapes.misc}
\usetikzlibrary{shadows}
\usetikzlibrary{arrows}
\usetikzlibrary{positioning}
\usetikzlibrary{matrix}
\usetikzlibrary{calc}
\usepackage{pgfplots}
\usepackage{wrapfig}
\usepackage{subcaption}

\newenvironment{figCustom}{%
	\begin{figure}
    }{%
    \end{figure}
    }

\setcitestyle{square}

\usepackage{axelPaper}

\begin{document}

\title{The Mechanical Neural Network(MNN) -- A physical implementation of a multilayer perceptron for education and hands-on experimentation}

\author{Axel Schaffland}
\orcid{0000-0002-4640-7864}
\affiliation{%
  \institution{Institute of Cognitive Science, University of Osnabr\"uck}
  \streetaddress{Wachsbleiche 25}
  \city{Osnabr\"uck}
  \state{NS}
  \postcode{49090}
  \country{GERMANY}}
\email{axschaffland@uni-osnabrueck.de}

\renewcommand\shortauthors{Schaffland, A.}

\keywords{Mechanical Computer, Analog Computer, Neural Network, Multilayer Perceptron, Teaching Material, Education}

\begin{abstract}
\glsresetall
In this paper the \gls{mnn} is introduced, a physical implementation of a 
\gls{mlp} with \gls{relu} activation functions, two input neurons, four hidden neurons
and two output neurons. 
This physical model of a \gls{mlp} is used in 
education to give a hands on experience and allow students to experience
the effect of changing the parameters of the network on the output. Neurons
are small wooden levers which are connected by threads. Students can adapt
the weights between the neurons by moving the clamps connecting a neuron 
via a thread to the next.  
The \gls{mnn} can model real valued functions and logical operators including XOR.

\end{abstract}

\begin{teaserfigure}
\includegraphics[width=\textwidth]{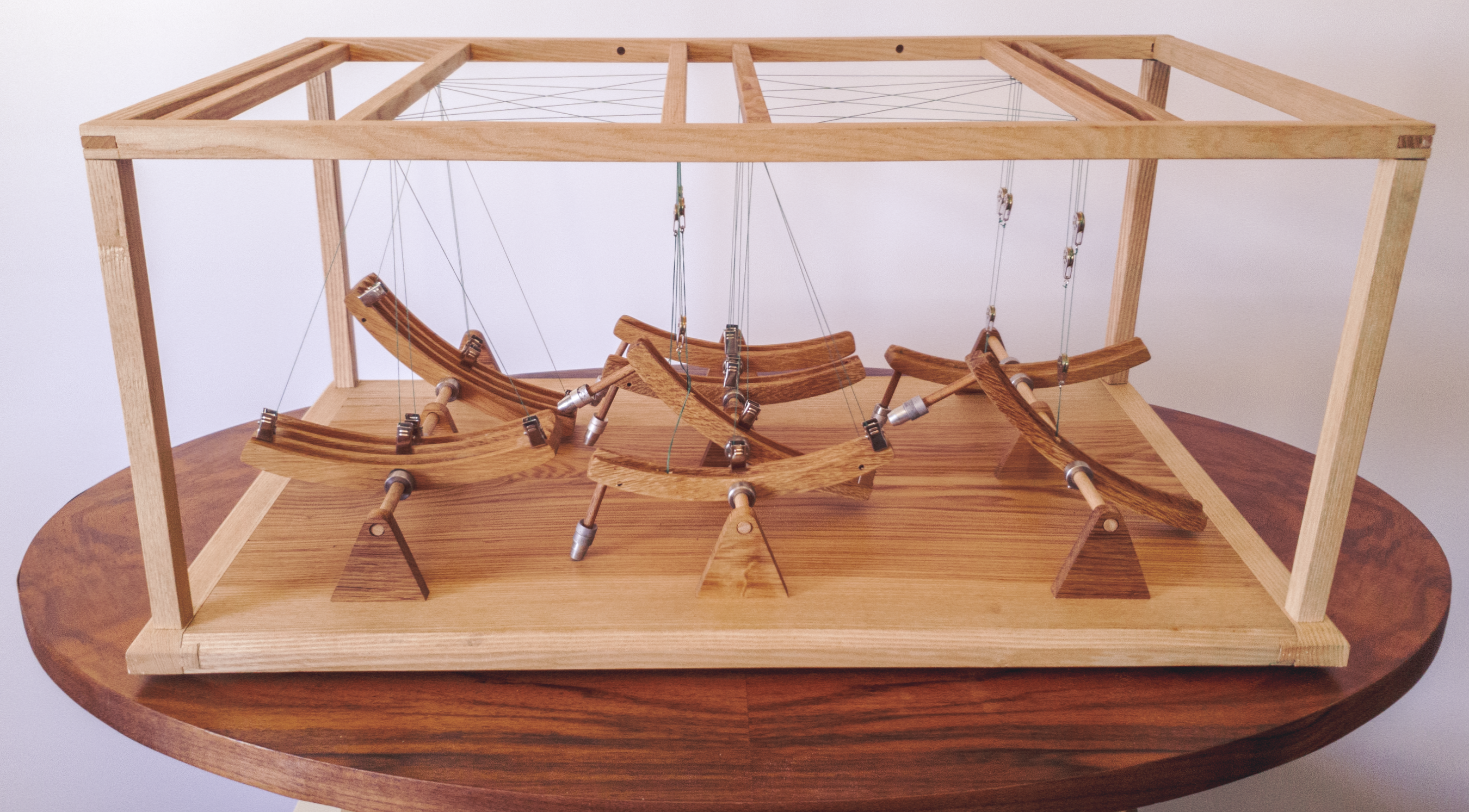}
\caption{The \gls{mnn} with the two input neurons on the left, the four hidden neuron in the middle and the two output neurons on the right. The strings are the connections between the neurons weighted by the position of the clamps on the sending neuron and combined by the pulley systems above the receiving neurons. A video can be found at \url{https://youtu.be/zMxh3Io3hFE}.}
\Description{A photograph of the wooden model. A wooden ground plane with a wooden frame above, seesaws and strings connecting the seesaws over the top of the frame.}
\label{fig:photo}
\end{teaserfigure}

\maketitle

\section{Introduction}
\glsresetall
\label{sec:intro}
The \gls{mnn}, presented in this paper, is a mechanical implementation
of a \gls{mlp} with \gls{relu} activation functions. Thus, it can model logical functions, including 
XOR but also real valued functions.
It is build from wooden levers, which represent the neurons. 
These levers are linked by strings, corresponding to the connections 
between the neurons of a \gls{mlp}. Adjusting the clamps, which connect 
the strings to the levers, by hand allows to adjust the weights of the 
network. Hence, the effect of adapting the weights can be intuitively observed.

Intuition for education is one of the motivating factors of the \gls{mnn}: Digital 
implementations of a \gls{mlp} or even more complex networks are
difficult to understand. The \gls{mnn} allows to develop this intuition
and facilitates the work with more complex models. 

Moreover, the \gls{mnn} shows that a small \gls{mlp} can be implemented purely
mechanical without the need for electricity and can be build to withstand harsh environments.

While many analog computers were developed\ci{ulmann2013,lundberg2005} most are electro-mechanical,
electrical, or electronic. Few are purely mechanical and none try to model a 
\gls{mlp}, especially for use in education. The fictional \emph{Rope and Pulley
Computer}, labelled as ancient discovery and published as April Fools' joke, 
models logical gates with ropes and pulleys but was never 
constructed\ci{dewdney1988}. The \emph{Tinkertoy Computer} uses ropes and 
construction kit hardware and was able to play Tic-tac-toe\ci{hillis1978},
however it is not field programmable / trainable like the \gls{mnn}. It is 
also complex and difficult to use for educational purposes, compared to the
\gls{mnn}. There exist at least on neural network model in a museum, a model visualizing
connections with fibre optics\ci{tmw2020}, but it can not be used in class nor 
does it mimic the functionalities of a neural network on a mechanical level. 

Analog computers today are not only relevant for education but are also sought
for harsh environments, in which normal computers fail to operate, \eg in a 
rover on the surface of Venus\ci{sauder2017a} (and with more details in\ci{sauder2017b}).
In principle a roughed (and miniaturized) version of the \gls{mnn} is able to work 
under those harsh conditions. 

The paper is organised in the following way: In Section\re{sec:technique}  
the mechanical structure of the \gls{mnn} is explained and it is shown that this
physical implementation models indeed a \gls{mlp}. Then, in 
Section\re{sec:problems}, some logical operators
are modelled with the \gls{mnn}. Next, it is shown how the \gls{mnn} can be used for teaching in Section\re{sec:education} and lastly further improvements to the network are presented in Section\re{sec:further}.

\section{The \glsentrylong{mnn} and why it models a \glsentrylong{mlp}}
\label{sec:technique}
\subsection{Notation}
In the following the $i$th neuron from layer $k$ will be denoted as $n_i^{(k)}$, its weighted input as $\text{net}_i^{(k)}$, and its output as $o_i^{(k)}$.
The weight from the $j$th neuron of the previous layer to the $i$th neuron of layer $k$ is denoted as $w_{ij}^{(k)}$.

\subsection{Architecture}
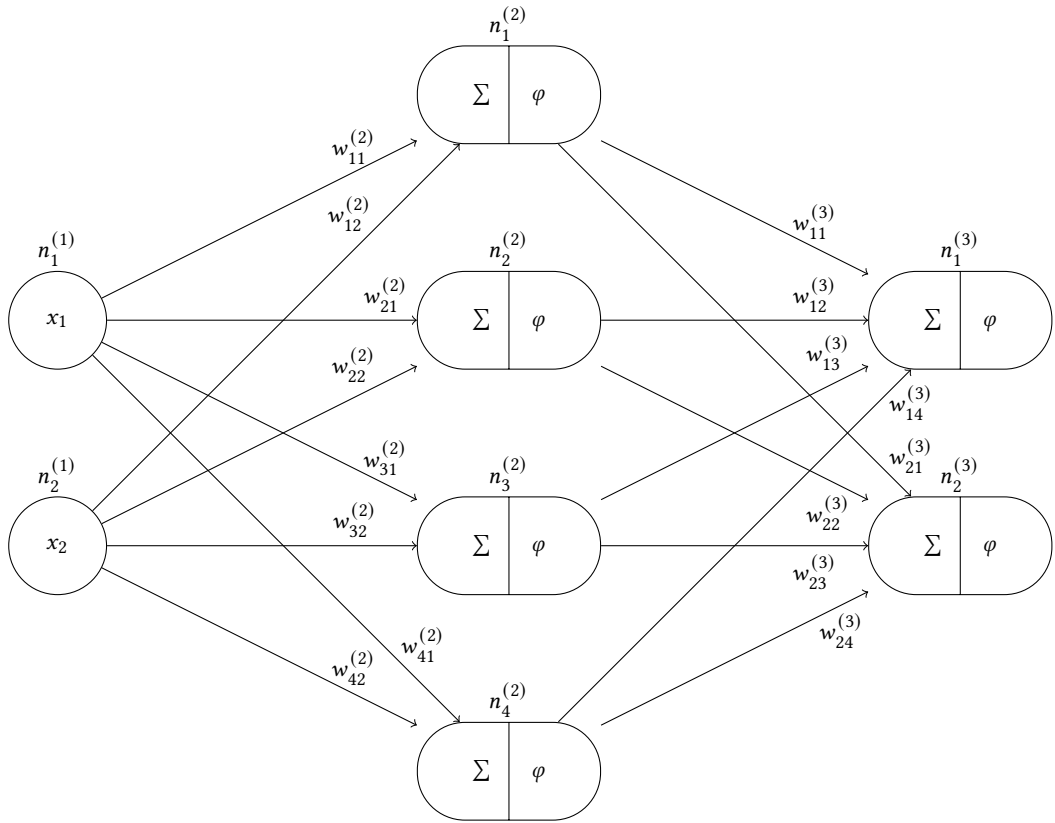
\begin{figure*}
	\begin{tikzpicture}[xscale=3,yscale=1.5]
    \tikzstyle{neuron}=[circle,draw=black,minimum size=37pt,inner sep=0pt]
    \tikzstyle{neuron2}=[%
                       rectangle split,
                       rectangle split horizontal,
                       rectangle split parts=2,
                       rounded corners=18pt,
                       minimum size=37pt,
                       minimum width=4.5cm,
                       text width=1cm,
                       draw=black]
    \node[neuron] (I1) at (0,5) {$x_1$};
    \node[above=-0.05 of I1]{$n_1^{(1)}$};
    \node[neuron] (I2) at (0,3) {$x_2$};
    \node[above=-0.05 of I2]{$n_2^{(1)}$};

     \node[neuron2] (H1) at (2,7) {
       \nodepart[text width=1cm]{one}
       \nodepart[text width=1cm]{two} } ;
     \node[above=-0.05 of H1]{$n_1^{(2)}$};
     \node at (H1) {$\sum$\hspace{4ex}$\varphi$} ;
     
     \node[neuron2] (H2) at (2,5) {
       \nodepart[text width=1cm]{one}
       \nodepart[text width=1cm]{two} } ;
     \node[above=-0.05 of H2]{$n_2^{(2)}$};
     \node at (H2) {$\sum$\hspace{4ex}$\varphi$} ;
     
     \node[neuron2] (H3) at (2,3) {
       \nodepart[text width=1cm]{one}
       \nodepart[text width=1cm]{two} } ;
     \node[above=-0.05 of H3]{$n_3^{(2)}$};
     \node at (H3) {$\sum$\hspace{4ex}$\varphi$} ;
    
    \node[neuron2] (H4) at (2,1) {
      \nodepart[text width=1cm]{one}
      \nodepart[text width=1cm]{two} } ;
     \node[above=-0.05 of H4]{$n_4^{(2)}$};
     \node at (H4) {$\sum$\hspace{4ex}$\varphi$} ;

     \node[neuron2] (O1) at (4,5) {
       \nodepart[text width=1cm]{one}
       \nodepart[text width=1cm]{two}} ;
     \node[above=-0.05 of O1]{$n_1^{(3)}$};
     \node at (O1) {$\sum$\hspace{4ex}$\varphi$} ;
     
     \node[neuron2] (O2) at (4,3) {
       \nodepart[text width=1cm]{one}
       \nodepart[text width=1cm]{two}} ;
     \node[above=-0.05 of O2]{$n_2^{(3)}$};
     \node at (O2) {$\sum$\hspace{4ex}$\varphi$} ;

    \draw[->] (I1) -- (H1) node[pos=.8,above] {$w_{11}^{(2)}$};
    \draw[->] (I1) -- (H2) node[pos=.9,above] {$w_{21}^{(2)}$};
    \draw[->] (I1) -- (H3) node[pos=.9,above] {$w_{31}^{(2)}$};
    \draw[->] (I1) -- (H4) node[pos=.9,above=.2] {$w_{41}^{(2)}$};
    
    \draw[->] (I2) -- (H1) node[pos=.7,above=.2] {$w_{12}^{(2)}$};
    \draw[->] (I2) -- (H2) node[pos=.8,above=.1] {$w_{22}^{(2)}$};
    \draw[->] (I2) -- (H3) node[pos=.8,above] {$w_{32}^{(2)}$};
    \draw[->] (I2) -- (H4) node[pos=.8,above] {$w_{42}^{(2)}$};

    \draw[->] (H1) -- (O1) node[pos=.8,above] {$w_{11}^{(3)}$};
    \draw[->] (H1) -- (O2) node[pos=1,above=.2] {$w_{21}^{(3)}$};
    
    \draw[->] (H2) -- (O1) node[pos=.8,above] {$w_{12}^{(3)}$};
    \draw[->] (H2) -- (O2) node[pos=.85,below=.1] {$w_{22}^{(3)}$};
    
    \draw[->] (H3) -- (O1) node[pos=.85,above=.1] {$w_{13}^{(3)}$};
    \draw[->] (H3) -- (O2) node[pos=.8,below] {$w_{23}^{(3)}$};
    
    \draw[->] (H4) -- (O1) node[pos=1,below=.1] {$w_{14}^{(3)}$};
    \draw[->] (H4) -- (O2) node[pos=.9,below] {$w_{24}^{(3)}$};

\end{tikzpicture}
	\caption{The architecture of the \gls{mnn} with one input (2 neurons), one hidden (4 neurons), and one output layer (2 neurons). Superscript indices indicate the layer and subscript indices the neuron. Weights are subscripted with the index of the receiving neuron followed by the index of the sending neuron and superscripted with the index of the receiving layer. The input of a neuron is denoted as $\sum$ to which the activation function $\varphi$ is applied.}
	\Description{Shown is the network architecture as graph with two nodes, representing the input neurons, on the left, four nodes in the middle, representing the hidden neurons, and two nodes on the right, representing the output neurons. The left nodes have edges to all middle nodes. All middle nodes have edges to all right nodes.}
	\label{fig:architecture}
\end{figure*}
The \gls{mnn} implements a \gls{mlp}\ci{rumelhart1986} with \gls{relu} activation functions and two input neurons, four hidden neurons and two output neurons as shown in Figure\re{fig:architecture}. The three layers are three wooden shafts through the fulcrums of the levers as shown in Figure\re{fig:photo}. The \gls{mnn} is a fully connected network. The connecting strings are clamped to the sending neuron, travel to the top of the model and are routed to a receiving neuron of the next layer. There they travel down into the pulley system, which combines the weighted inputs of this receiving neuron. A single string then exists the pulley system, the weighted sum, and connects to the receiving neuron.
\subsection{Neurons}
\begin{figCustom}
	\center
	\includegraphics[width=.8\textwidth]{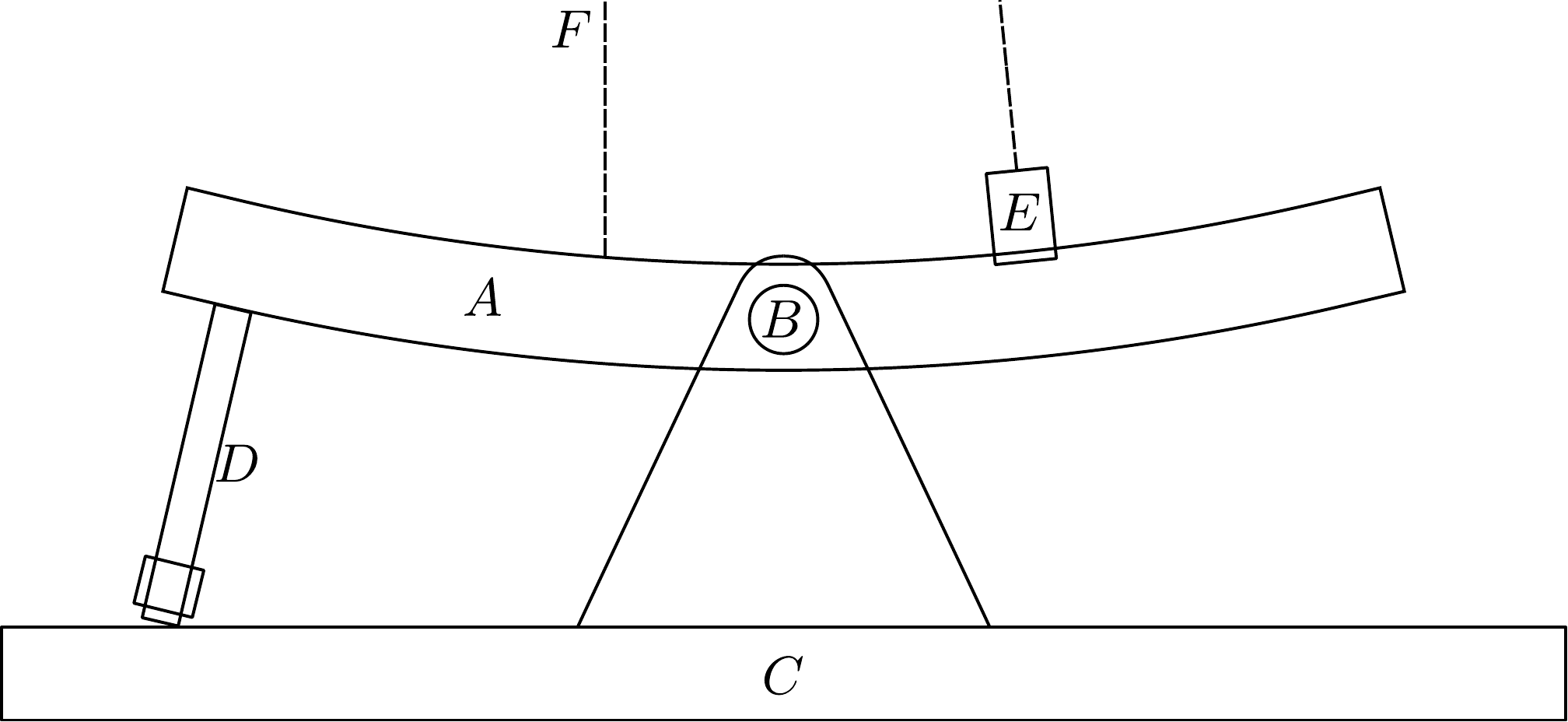}	
	\caption{A single lever of the \gls{mnn} representing a neuron. The lever $A$ rotates around its fulcrum $B$. Clockwise it rotates until it contacts the ground plane $C$. Counterclockwise rotation is prohibited by the weighted rod $D$ both acting as counterweight, resetting the lever to the horizontal position, and acting as \gls{relu} activation function. The clamp $E$ can be moved along the top of the lever and connects via a string to a lever in the next layer. Moving the clamp corresponds to adjusting the weight. $F$ is the string corresponding to the input of this neuron. Note that there is a clamp and connecting string for each neuron in the next layer. Also note that for the input neurons string $F$ and the rod $D$ is missing, allowing these levers to rotate counterclockwise, i.e. the input neurons have no activation function.}
	\Description{A technical drawing of a arced seesaw with a string clamped to it leading to the center of the arc.}
	\label{fig:neuronLever}
\end{figCustom}
Each of the neurons is a lever (Figure\re{fig:neuronLever}) rotating around a shaft at its fulcrum. Each of the three shafts corresponds to one of the three layers. Horizontal levers correspond to neurons with output $o=0$, levers rotated clockwise correspond to neurons with $o>0$ and counterclockwise rotated levers to neurons with $o<0$. Rotating the neurons till they contact the ground plane is defined as $-1$,$+1$ respectively, i.e. $o \in [-1,+1]$ (not taking the activation function into account). 

Each neuron is connected to all neurons of the next layer by strings clamped to this neuron. Clamps can be moved on the lever of the sending neuron, representing different weights on the connection from this neuron to the receiving neuron in the next layer. A clamp in the middle at the pivot of the lever is equivalent to $w=0$, since the connecting string is not moved when the lever is moved. A clamp on the right side of the lever corresponds to $w>0$ and the receiving lever rotates in the same direction while a clamp on the left side of the lever correspond to $w<0$ and the receiving neurons rotates in the opposite direction. Moving the clamp to the full left or right corresponds to $w=1$,$w=0$ respectively. Hence, weights are limited to the range $[-1,+1]$.

Adjusting the weights, \ie moving the clamps, does not change the bias of the receiving neurons, since the levers, on which the clamps are moved, are arcs with the centre being the deflection eye of the string at the top of the model. 

\subsection{Layers}
Each layer is a shaft around which the neurons of that layer can rotate. The two input neurons can be rotated by hand around the shaft, stay in position and represent the input. As in a \gls{mlp} the input neurons do not have an activation function and represent the input to the network. 

The input layer is followed by a hidden layer with four neurons and an output layer with two neurons. Note that for the two-valued logical operators, discussed below, two hidden and one output neuron suffice.

The \gls{mnn} is fully connected. This means that each lever in the input layer has four clamps and four strings connecting it to each of the four levers in the hidden layer. Thus, the four levers of the hidden layer have two clamps and strings each, connecting them to the two levers of the output layer.

\subsection{Weights and Net Input}
\begin{figCustom}
	\center
	\includegraphics[width=.2\textwidth]{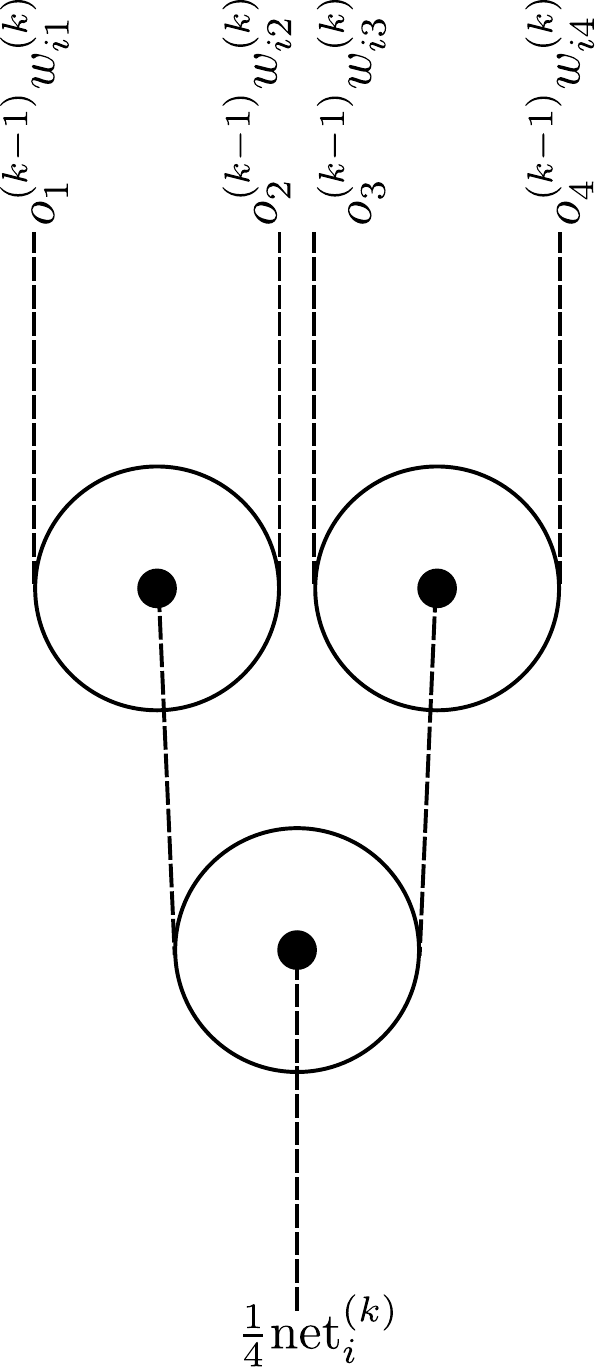}	
	\caption{The pulley system to compute the sum of the weighted inputs of neuron $i$ in the output layer. At the top the four strings from the clamps on the four levers of the hidden layer enter. Their movement corresponds to the weighted inputs of neuron $i$. The movement is combined/summed by a two stage pulley system resulting in the net input for neuron $i$. Note that the movement is reduced by a factor of $\frac{1}{4}$ by the pulley system. To address this the string, representing $\frac{1}{4}\text{net}_i^{(k)}$ is connected to the lever $i$ not at the most left but at $\frac{1}{4}$ of the distance between the fulcrum and the most left, resulting in the correct input $\text{net}_i^{(k)}$.}
	\Description{A drawing of a pulley system with 3 pulleys, 4 strings at the top looping around two pulleys, a string connecting these pulleys and looping around the third pulley below and a string connected at the bottom connected to the center of the third pulley.}
	\label{fig:weightPulley}
\end{figCustom}

For a single perceptron\ci{rosenblatt1958} the net input of a neuron is defined as
$$
net^{(k)}_i = \underset{j}{\sum}o^{(k-1)}_j  w^{(k)}_{ij} + b^{(k)}_i
$$
with $b^{(k)}_i$ being the bias of that neuron. In the \gls{mnn} the net input is simplified to a version without bias. However, note that it can be reintroduced by setting the output of a neuron from the previous layer to $1$ and interpreting the corresponding weight as bias. In the model this corresponds to rotating the corresponding lever fully clockwise and using the position of the clamps on this lever as biases. (This is analogous to the vector notation of the net input: $\text{net}_i^{(k)} = \vec{o}^{(k-1)^T}\vec{w}_i^{(k)}$ with $o^{(k-1)}_0=1$.)

This weighted sum is realized in two steps in the \gls{mnn}: The weights are set with the clamps as described above. The rotation of the neuron translates to a linear movement of the connecting string to a neuron in the next layer. The position of the clamp governs the transmission ratio from rotational movement of the lever to the linear movement of the string, \ie weights the output of this neuron. 
(The number of clamps on this neuron is equal to the number of neurons in the next layer.)

The summation is done with a pulley system as described in Figure\re{fig:weightPulley}. The pulley system is connected to the neuron in the next layer by another string, translating the linear movement of the pulley system to rotational movement. I.e. the rotation of this lever represents the input and output of that neuron, unless the rotation is limited by the \gls{relu} activation function. If that is the case the string becomes slack and the rotation represents only the output of the neuron.

\subsection{Activation Function}
Neurons of the hidden and the output layer use the \gls{dorelu} activation function, based on the \gls{relu} activation function\ci{fukushima1969}. They are defined $\varphi(x)=\min(1,\max(0,x))$ -- a plot is shown in Figure\re{fig:relu}. This activation function is easily modelled in the \gls{mnn} by restricting the rotation of the lever to clockwise rotations, since these correspond to positive output of the neuron, and prevent counter-clockwise rotations, since these correspond to negative output to the neuron. This restriction is implemented as a vertical bar \textit{D} connected to the lever \textit{A} and resting on the ground plane \textit{C} and thus preventing counter-clockwise rotation (Compare Figure\re{fig:neuronLever}). Clockwise rotation in limited by the lever contacting the ground plane, corresponding to an activation of $+1$. Thus, for the neurons in the hidden and the output layer: $o \in [0,+1]$.

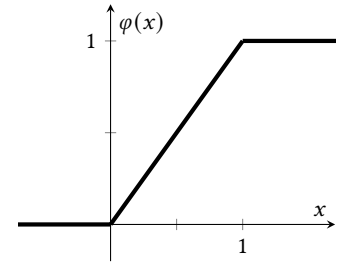
\begin{wrapfigure}{rl}{.3\textwidth}
	\centering
	\begin{tikzpicture}                                                              
  \begin{axis}[axis lines=middle,height=5cm,                                     
    clip=true,                                                                   
    xmin=-.7,                                                                    
    xmax=1.7,                                                                     
    ymax=1.2,                                                                     
    ymin=-.2,      
    xtick={.5,1}, 
    ytick={.5,1},                                                            
    xticklabels={,1},                                                              
    yticklabels={,1},                                                              
    domain=-.8:1.6,smooth,                                                        
    samples=50,                                                                  
    xlabel={$x$},                                                                
    ylabel={$\varphi(x)$},                                                       
    ]                                                                                                                               
    \addplot[ultra thick,mark=none,domain=-1:0] {0};                      
    \addplot[ultra thick,mark=none,domain=0:1] {x};                                                                                            
    \addplot[ultra thick,mark=none,domain=1:2] {1};                                                                                            
  \end{axis}                                                                     
\end{tikzpicture}  	
	\caption{Plot of the double \gls{relu} activation function $\varphi(x)$.}
	\label{fig:relu}
	\vspace{-1cm}
\end{wrapfigure}

\subsection{Differences between \gls{mlp} and \gls{mnn}}
The \gls{mnn} differs in four points from a \gls{mlp}. The \gls{mnn} has (1) no bias, (2) weights limited to the interval$[-1,1]$, (3) inputs limited to the interval $[-1,1]$, and (4) \gls{dorelu} activation functions. (1) can be addressed by setting a neuron to one and using its weights as bias, (2),(3) and (4) could be addressed by input scaling. However, this is not relevant for two-valued logical operators and educational purposes.

\section{Modelling exemplary functions}
\label{sec:problems}
\subsection{Logical Operators}
\begin{wraptable}{lr}{.4\textwidth}
\centering
\begin{tabular}{c|c|c|c|c|c}
	$x_1$	 						&	$x_2$ &		NOT($x_1$) &		AND &		OR  & XOR\\
	\hline
	0 & 0 & 1 & 0 & 0 & 0\\
	0 & 1 & 1 & 0 & 1 & 1\\
	1 & 0 & 0 & 0 & 1 & 1\\
	1 & 1 &	0 & 1 & 1 & 0\\	
\end{tabular}
\caption{The truth table of the modelled logical operations}
\label{tab:truthTab}
\vspace{-1cm}
\end{wraptable}

For boolean algebra a horizontal lever is interpreted as $false$ and a lever fully rotated clockwise as $true$. The truth table for the modelled logic operations is Table\re{tab:truthTab}.

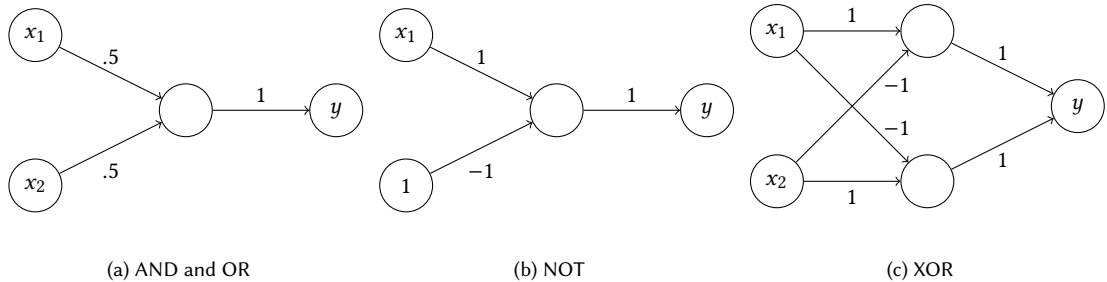
\begin{figure}[b]
    \centering
    \begin{subfigure}[b]{0.3\textwidth}
	\begin{tikzpicture}[xscale=1,yscale=1]
    \tikzstyle{neuron}=[circle,draw=black,minimum size=20pt,inner sep=0pt]
    \tikzstyle{neuron2}=[%
                       rectangle split,
                       rectangle split horizontal,
                       rectangle split parts=2,
                       rounded corners=18pt,
                       minimum size=37pt,
                       minimum width=4.5cm,
                       text width=1cm,
                       draw=black]
    \node[neuron] (I1) at (0,2) {$x_1$};
    \node[above=-0.05 of I1]{};
    \node[neuron] (I2) at (0,0) {$x_2$};
    \node[above=-0.05 of I2]{};

     \node[neuron] (H1) at (2,1) {} ;
     \node[above=-0.05 of H1]{};       

     \node[neuron] (O1) at (4,1) {$y$} ;
     \node[above=-0.05 of O1]{};

    \draw[->] (I1) -- (H1) node[pos=.5,above] {$.5$};

    \draw[->] (I2) -- (H1) node[pos=.5,below=.1] {$.5$};

    \draw[->] (H1) -- (O1) node[pos=.5,above] {$1$};

\end{tikzpicture}	
	\caption{AND and OR}
	\Description{A graph with 2 nodes on the left, 2 in the middle and 1 one the right with edges between the nodes to the left and the middle and the nodes in the middle and the right node}
	\label{fig:andor}
    \end{subfigure}
    \quad %
    \begin{subfigure}[b]{0.3\textwidth}
	\begin{tikzpicture}[xscale=1,yscale=1]
    \tikzstyle{neuron}=[circle,draw=black,minimum size=20pt,inner sep=0pt]
    \tikzstyle{neuron2}=[%
                       rectangle split,
                       rectangle split horizontal,
                       rectangle split parts=2,
                       rounded corners=18pt,
                       minimum size=37pt,
                       minimum width=4.5cm,
                       text width=1cm,
                       draw=black]
    \node[neuron] (I1) at (0,2) {$x_1$};
    \node[above=-0.05 of I1]{};
    \node[neuron] (I2) at (0,0) {$1$};
    \node[above=-0.05 of I2]{};

     \node[neuron] (H1) at (2,1) {} ;
     \node[above=-0.05 of H1]{};       

     \node[neuron] (O1) at (4,1) {$y$} ;
     \node[above=-0.05 of O1]{};

    \draw[->] (I1) -- (H1) node[pos=.5,above] {$1$};

    \draw[->] (I2) -- (H1) node[pos=.5,below=.1] {$-1$};

    \draw[->] (H1) -- (O1) node[pos=.5,above] {$1$};

\end{tikzpicture}	
	\caption{NOT}
	\Description{A graph with 2 nodes on the left, 2 in the middle and 1 one the right with edges between the nodes to the left and the middle and the nodes in the middle and the right node}
	\label{fig:not}
    \end{subfigure}
    \quad%
    \begin{subfigure}[b]{0.3\textwidth}
	\begin{tikzpicture}[xscale=1,yscale=1]
    \tikzstyle{neuron}=[circle,draw=black,minimum size=20pt,inner sep=0pt]
    \tikzstyle{neuron2}=[%
                       rectangle split,
                       rectangle split horizontal,
                       rectangle split parts=2,
                       rounded corners=18pt,
                       minimum size=37pt,
                       minimum width=4.5cm,
                       text width=1cm,
                       draw=black]
    \node[neuron] (I1) at (0,2) {$x_1$};
    \node[above=-0.05 of I1]{};
    \node[neuron] (I2) at (0,0) {$x_2$};
    \node[above=-0.05 of I2]{};

     \node[neuron] (H1) at (2,2) {} ;
     \node[above=-0.05 of H1]{};
     
     \node[neuron] (H2) at (2,0) {} ;
     \node[above=-0.05 of H2]{};

     \node[neuron] (O1) at (4,1) {$y$} ;
     \node[above=-0.05 of O1]{};

    \draw[->] (I1) -- (H1) node[pos=.5,above] {$1$};
    \draw[->] (I1) -- (H2) node[pos=.9,above=.1] {$-1$};

    \draw[->] (I2) -- (H1) node[pos=.9,below=.1] {$-1$};
    \draw[->] (I2) -- (H2) node[pos=.5,below] {$1$};

    \draw[->] (H1) -- (O1) node[pos=.5,above] {$1$};

    \draw[->] (H2) -- (O1) node[pos=.5,below] {$1$};

\end{tikzpicture}	
	\caption{XOR}
	\Description{A graph with 2 nodes on the left, 2 in the middle and 1 one the right with edges between the nodes to the left and the middle and the nodes in the middle and the right node}
	\label{fig:xor}
    \end{subfigure}
    \caption{The \gls{mnn} configurations for different logical operators}\label{fig:logicalConfigurations}
\end{figure}

\subsubsection{AND and OR}
For the logical operators AND and OR only two inputs, one hidden and one output neuron is used as show in Figure\re{fig:andor}. The implementation differs only in the threshold applied to the output of the \gls{mnn}: For OR, $y\geq0.5$ would evaluate to $true$ and for AND, $y\geq1$ would evaluate to $true$. Another solution is to introduce a bias and threshold at $0$. This relates to the fact that the separatrixes for AND and OR differ only by the intercept.

\subsubsection{NOT}
The NOT operation takes only one input and the output is the inverse of the input. Here the second input neuron is set to $1$ and its weight is used as bias as shown in Figure\re{fig:not}. 

\subsubsection{XOR}
For the XOR operation the output is $true$ iff one of the two inputs is $true$ and $false$ otherwise. Here a symmetric network configuration with two hidden neurons is used, shown in Figure\re{fig:xor}. One input activates one hidden neuron and inhibits the other. This example highlights the importance of the nonlinear activation functions: Without the activation functions the output of the network would always be zero. 
While the previous logical operations are linearly separable and can also be modelled with a perceptron, this is not the case for the XOR\ci{minsky2017} -- it is not linearly separable and requires an \gls{mlp} with nonlinear activation functions.

\section{Application in Education}
\label{sec:education}
The \gls{mnn} is best used in a hands-on approach in which students can design and adapt the network themselves, i.e. adjust levers and clamps themselves, after they where familiarized with its functionality. Although, the \gls{mnn}  can be trained for real valued problems, logic operators are better suited for education, as they are easier to understand and require less 
fineness when adjusting the weights.

In a first stage students may freely experiment to adapt the \gls{mnn} to different logical operators. A good starting point are the  AND and OR operators. Here students can discover that both operators are linearly separable, that the non-linear hidden layer is not needed, and that the two operators vary only in a different threshold on the output (or a different bias of the linear separator). A thresholded sum of the two inputs suffices and the \gls{mnn} can indeed be used to sum boolean (and real valued) inputs. 

Next students can adapt the \gls{mnn} for the XOR operator. They can discover that this problem is not linear separable and needs indeed a non-linear hidden layer. They can also discover that this symmetric problem can be solved by a symmetric network configuration as shown in Figure\re{fig:xor}. Further, the analogy between neurons and logical gates and the possibility to express logic operators as combination of other logic operators can be discovered. \Eg the configuration of the \gls{mnn} for the XOR operator stated above is equal to $(x_1 \land \neg x_2) \lor (\neg x_1 \land x_2)$. From this students can observe, that the input space, in which the XOR operator is not linearly separable, is transformed into the space spanned by $(x_1 \land \neg x_2)$ and $(\neg x_1 \land x_2)$ by the hidden neurons. In this space the operator is then linearly separable by  the output neuron.

With the introduction of the NOT operator students will discover that a bias is needed.  It can be represented by completely activating one of the input neurons and using its weights as bias, linking to the vector notation of neurons.

In the first stage students will  discover that setting the correct weights by hand and intuition can be difficult, leading to a second stage and raising the question if a procedure exists how the network can be trained by hand and motivating automatic training algorithms like backpropagation. The concept of an error between desired and actual output can be introduced, as well as stochastic gradient descent. 

This would be a good point to switch over to a computer simulation of the \gls{mnn}, and then extend to implementations of \glspl{mlp} and later deep learning frameworks.

\section{Further Work}
\label{sec:further}
The \gls{mnn} can be further developed in different directions: (1) For classroom use several models would be useful, hence more economic (and also more robust) versions are desirable -- possibly with 3D printed parts and wooden strips of standardized size. (2) For use in harsh environments a roughed, miniaturized, and encapsulated version could be development, using plastic and metal components and springs instead of counterweights to keep levers in position. Here, more neurons and layers would also seem desirable. (3) Development of a process to train the network by hand, guiding which weights are to be adapted to generate a specific output. (4) Complete simulation of the \gls{mnn} with an adapted backpropagation algorithm. The \gls{mnn} can then be trained in the simulation and the resulting weights can be set in the real model. This is especially useful for real valued classification and regression problems for which the optimal weights can not be found intuitively by hand. 

\section{Conclusion}
\label{sec:conclusion}
In this paper the \gls{mnn}, a mechanical model of a \gls{mlp} was presented. It consists of an input layer with two neurons, a hidden layer with four neurons and an output layer with two neurons. However, other and more complex configuration are possible, too. Neurons are wooden levers which can be rotated, whereby the rotation of the lever represents the activation of the neuron. The levers are connected with strings, passing the information from a sending neuron to the receiving neurons in the next layer. Albeit, the \gls{mnn} is not limited to logical operators, they are easy to understand and training the network for this operators can be done intuitively by hand. This allows to use the \gls{mnn} in educational settings, in which students experiment and model these different logical operators, allowing to gain insights into a \gls{mlp} and motivating training algorithms like backpropagation. While the \gls{mnn} can also be implemented as a roughed mechanical computer suitable for harsh environments, its current main focus is on education in a museum and especially classroom environment.

\bibliographystyle{ACM-Reference-Format}
\bibliography{../rePhotos.bib}

\end{document}